# Problems and shortcuts in deep learning for screening mammography

Trevor Tsue, Brent Mombourquette, Ahmed Taha, Thomas Paul Matthews, Yen Nhi Truong Vu, Jason Su

## Abstract

### Background

Deep learning models may exploit spurious shortcuts during training. Understanding performance and correcting shortcuts are essential for ensuring safety and reliability in clinical use.

### Purpose

This work reveals undiscovered challenges in the performance and generalizability of deep learning models. We (1) identify spurious shortcuts and evaluation issues that can inflate performance and (2) propose training and analysis methods to address them.

### Materials and methods

We trained an AI model to classify cancer on a retrospective dataset of 120,112 US exams (3,467 cancers) acquired from 2008 to 2017 and 16,693 UK exams (5,655 cancers) acquired from 2011 to 2015. We presented methods to determine whether an attribute is spuriously correlated with cancer: cancer prevalence, model output probability distributions, bias-aligned vs bias-conflicting performance, and attribute prediction. We applied these methods to four shortcuts: view markers, dataset, exam type, and scanner model. We mitigated these shortcuts by removing view markers, dataset balancing, and removing diagnostic exams. We evaluated performance using the area under the receiver operating characteristic curve (AUC) with a bootstrapped 95% confidence interval (CI).

### Results

We evaluated on a screening mammography test set of 11,593 US exams (102 cancers; 7,594 women; age 57.1 ± 11.0) and 1,880 UK exams (590 cancers; 1,745 women; age 63.3 ± 7.2). A model trained on images of only view markers (no breast) achieved a 0.691 [95% CI: 0.644, 0.736] AUC. The original model trained on both datasets achieved a 0.945 [0.936, 0.954] AUC on the combined US+UK dataset but paradoxically only 0.838 [0.792, 0.878] and 0.892 [0.875, 0.906] on the US and UK datasets, respectively. Sampling cancers equally from both datasets during training mitigated this shortcut. A similar AUC paradox (0.903 [0.886, 0.919]) occurred when evaluating diagnostic exams vs screening exams (0.862 [0.837, 0.884] vs 0.861 [0.818 0.898], respectively). Removing diagnostic exams during training alleviated this bias. Finally, the model did not exhibit the AUC paradox over scanner models but still exhibited a bias toward Selenia Dimension (SD) over Hologic Selenia (HS) exams. Analysis showed that this AUC paradox occurred when a dataset attribute had values with a higher cancer prevalence (dataset bias) and the model consequently assigned a higher probability to these attribute values (model bias). Stratification and balancing cancer prevalence can mitigate shortcuts during evaluation.

# Conclusion

Dataset and model bias can introduce shortcuts and the AUC paradox, potentially pervasive issues within the healthcare AI space. Our methods can verify and mitigate shortcuts while providing a clear understanding of performance.

# Summary Statement

Deep learning models in screening mammography can learn spurious shortcuts to artificially inflate model performance; stratification and balancing cancer prevalence can mitigate shortcut learning and elicit true model performance.

# Key results

- This paper introduces methods to verify the existence of shortcuts and identifies four shortcuts in mammography.
- We introduce methods to mitigate the effects of these shortcuts.
- We introduce the AUC paradox, showcasing how cancer prevalence, model bias, and dataset composition can inflate or deflate model performance.
- We emphasize the importance of stratification and balancing cancer prevalence during evaluation to understand model performance.

# 1 Introduction

A shortcut is the exploitation of dataset attributes spuriously correlated with the task.[1] Recent literature has shown that deep learning (DL) models can learn shortcuts instead of actual signals. Lapuschkin et al. (2019) identified that neural networks used signals from source tags and artificial padding.[2] Geirhos et al. (2020) proposed shortcut learning, in which models learned features from standard benchmarks but failed in real-world scenarios.[3] Carter et al. (2021) showed that models can confidently predict by overinterpreting pixels that are nonsensical to humans.[4]

In radiology, shortcuts are also of growing interest. Zech et al. (2018) showed that models classifying pneumonia could identify the hospital system and department while failing to generalize to three external sites.[5] Jabbour et al. (2020) demonstrated that DL models learned patient attributes like age, sex, and body mass index from a chest X-Ray by correlating them with diseases. Luo et al. (2021) illustrated that CheXNet gradCAM highlighted artifacts and shortcuts in chest radiographs.[6] DeGrave et al. (2021) revealed that artificial intelligence (AI) models identified the presence of Covid-19 from the data source, since the positive and negative images came exclusively from different datasets.[7] Gichoya et al. (2022) showed that DL models could identify race in chest x-rays whereas human experts could not.[8] These papers argue that models can learn shortcuts that inhibit generalization to new clinics.

DL algorithms have been utilized to identify breast cancer in mammograms for different applications: computer-aided detection[9,10], second reader automation[9,11,12], triage[9,11], and ruling out of non-cancer exams[13–19]. To work in clinical practice, these models must learn features that generalize. However, these models can exploit spurious correlations that obscure performance on different stratifications.[20] Thus, shortcuts threaten safe adoption and prevent greater clinical utility, a challenge not yet addressed by the radiology community.

This study focused on four objectives related to shortcuts in screening mammography. Our first objective is to present methods to identify spurious correlations exploited for malignancy classification (e.g., view markers, dataset source, exam type, and scanner model). Our second objective is to propose methods to mitigate these shortcuts during training, including removing view markers, balanced sampling from each dataset, and removing diagnostic exams. Our third objective is to demonstrate how combining datasets can result in an AUC greater than those of the individual datasets, which we term the AUC paradox. Our fourth objective is to highlight how the use of stratification can prevent this obfuscation of results.

# 2 Materials and Methods

This study used anonymized and retrospectively collected screening and diagnostic mammography exams. It was approved by the relevant institutional review boards (IRBs), and patient consent was waived for use of the anonymized data.

## 2.1 Data

Full-field digital mammography (FFDM) exams were collected from two institutions: Washington University in St. Louis (WUSTL) in the US, acquired from 2008 to 2017, and the UK National Health Service OPTIMAM (OPTIMAM),[21] acquired from 2011 to 2015. WUSTL exams were interpreted by 11 radiologists specializing in breast imaging ranging from 2 to 30 years of experience. OPTIMAM exams were interpreted by 213 radiologists of varying experience. Figure 1a describes the exclusion criteria for exams. We divided each dataset into three non-overlapping sets at the patient level: 80% train, 10% val and 10% test. Finally, we used the procedure in Figure 1b to establish ground truth for malignancy using a 12-month biopsy window. Table 1 breaks down the demographic characteristics of each exam. Table 2 reports the statistics and cancer prevalence of these datasets by malignancy label stratified over the exam type, screening vs diagnostic, and scanner model, Hologic Selenia (HS) vs Selenia Dimensions (SD). Additional data information is provided in Supplementary Section 1.

## 2.2 DL Model

Models were trained on both screening and diagnostic images from WUSTL and OPTIMAM to classify an FFDM image as cancer or non-cancer. All models employ the EfficientNet-B0[22] architecture implemented in PyTorch[23]. Initialized from pre-trained ImageNet[24] weights, the model was first trained on 512x512 patches centered on the bounding box for cancers or on random breast regions for non-cancers. The model was then trained on the full images, resized to 1664x1280. The image scores for a given breast were averaged to create a breast score, and the maximum of the breast scores was taken as the exam score (Supplementary Section 2).

We employed an iterative procedure to identify and mitigate shortcuts. Once a mitigation strategy was identified, this strategy was leveraged for all subsequent models. The original model removed all view markers and artifacts outside of the breast in the input image. The dataset-balanced model additionally sampled cancers and non-cancers equally from WUSTL and OPTIMAM. Finally, the dataset-balanced-screen-only model removed diagnostic images and only used screening images during training.

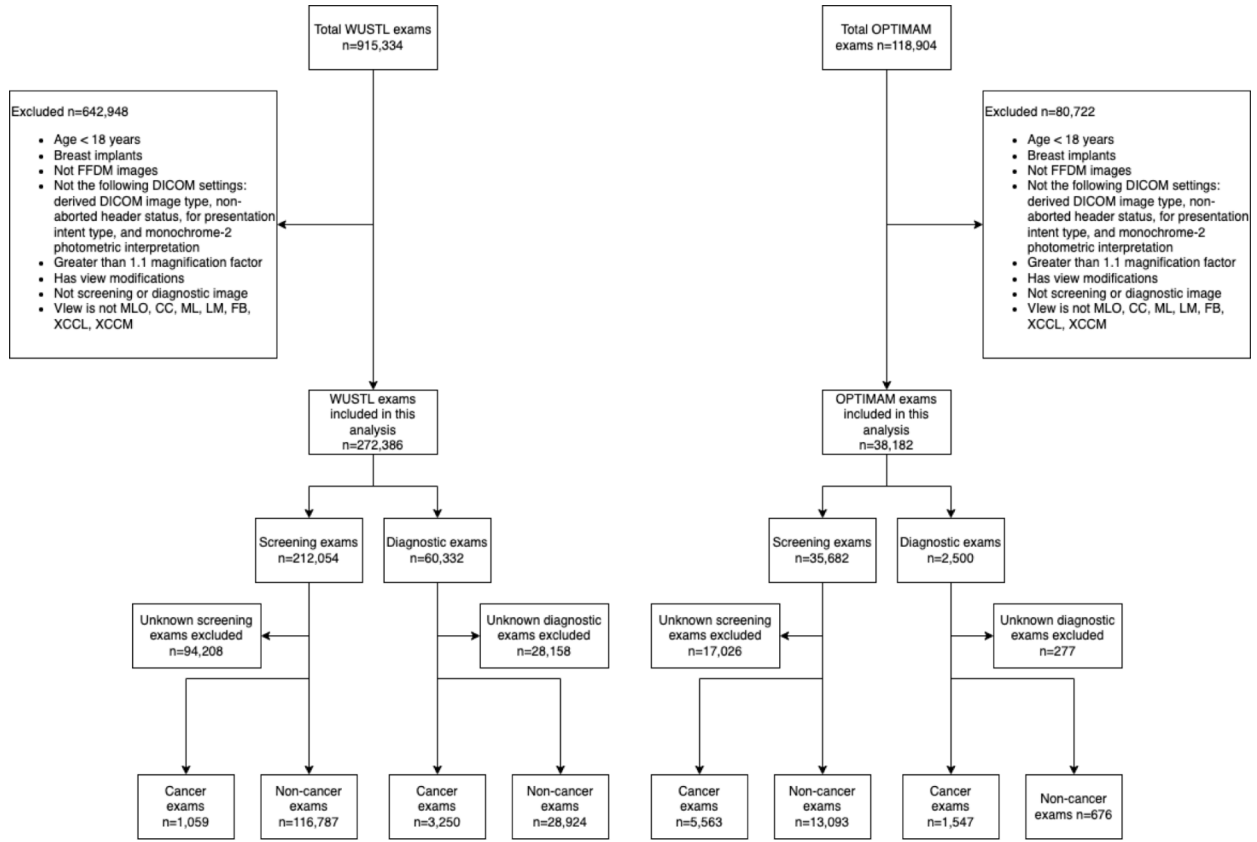

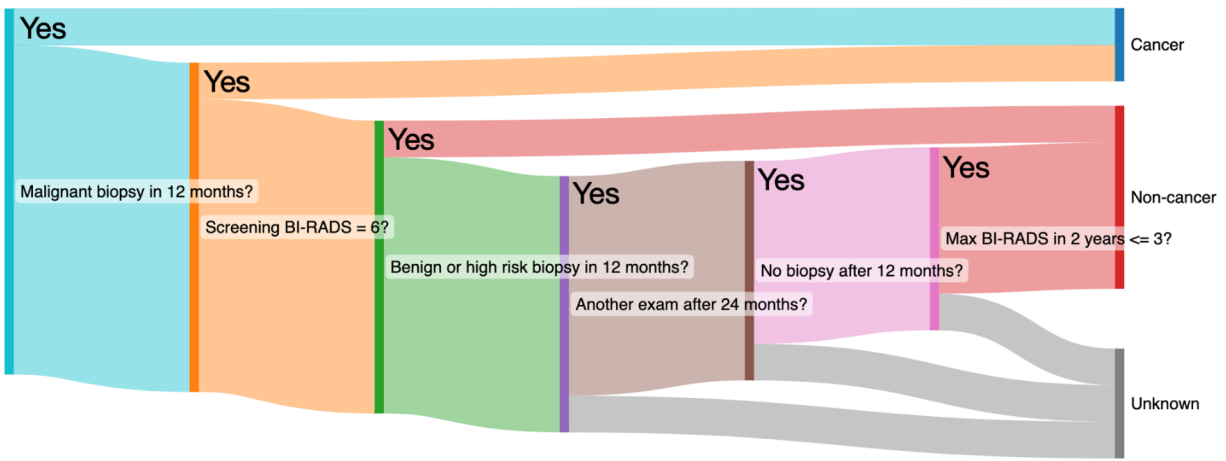

Figure 1: (a) Breakdown of the inclusion and exclusion criteria and their impact on the cases analyzed for the WUSTL and OPTIMAM datasets. (b) Ground truth labeling procedures for cancer and non-cancer mammography exams (width not to scale). An exam is labeled as cancer if there is a malignant biopsy within 12 months (the standard US screening interval) or if the exam receives a BI-RADS 6 label. This definition includes cancers found between screening exams, known as interval cancers. If there is no future biopsy after 12 months and there are two years of non-cancer follow-up with BI-RADS less than or equal to 3, the exam is labeled as non-cancer. These non-cancers include benign assessments found in screening exams, diagnostic exams, and benign or high-risk biopsies within 12 months. The remaining exams are labeled as unknown and are not used for training, validation, or testing.

## 2.3 Overview experimental approach

We introduce two types of biases for shortcut learning: (1) dataset bias and (2) model bias. Dataset bias occurs when a dataset attribute has multiple values with some having a higher cancer prevalence than others. Model bias occurs when the model learns to assign a higher probability to values with that higher cancer prevalence.

To verify dataset bias, we examined how cancer prevalence varied for different attribute values. To investigate model bias, we compared the average probability from model output distributions stratified by attribute values (e.g., US vs UK). We also split the test set into two pathological subsets along the attribute to compare performance on data that aligns and conflicts with model bias: (1) the bias-aligned and (2) bias-conflicting subsets. Namely, the bias-aligned subset has cancers with the high cancer prevalence attribute values and non-cancers with the low cancer prevalence attribute values (e.g., UK cancers and US non-cancers). The complementary bias-conflicting subset takes cancers with the low cancer prevalence attribute values and non-cancers with the high cancer prevalence attribute values (e.g., UK non-cancers and US cancers). Next, we train both (1) DL models on FFDM images and (2) logistic regression (LR) models on the malignancy model pre-logit feature vector to directly predict the values of the attributes (Supplementary Section 3). Intuitively, the ability to predict attribute values reveals whether attribute information is present in the (1) image or (2) malignancy model pre-logit feature vector.

Through our investigation of shortcuts, we observe the AUC paradox, a manifestation of Simpson's paradox[25] in which the AUC of a combined dataset can be greater than that of either of its individual component datasets. To understand this paradox, we introduced a simulation study to mimic dataset and model bias. We constructed two datasets with the exact same target AUC, modeling the positive and negative predictions as two separate normal distributions[26], tuning the mean and variance to control the individual target AUC (Supplementary Section 4). We investigated how altering the positive prevalence (dataset bias) and model bias of the second dataset affected the combined AUC.

## 2.4 Statistical Methodology

The main method of evaluation was comparison of the area under the receiver operating characteristic curve (AUC). To compare probability distributions between attributes, we used the Kolmogorov-Smirnov[27] (KS) statistic. The 95% confidence intervals (CI) were computed using percentile bootstrapping with replacement for 10,000 bootstrap replicates. All calculations used NumPy[28] version 1.19.0 and SciPy[29] version 1.5.4.

# 3 Results

## 3.1 Study Population

Table 1: Study populations for the training (train), validation (val), and test sets for WUSTL and OPTIMAM. The populations are stratified by exam type, number of unique women, age, race, and BI-RADS breast density. For age, race, and BI-RADs density, the exam counts are given. N/A means information is not available

| **Characteristic** | **WUSTL** | | | **OPTIMAM** | | |
|---|---|---|---|---|---|---|
| | **Train** | **Val** | **Test** | **Train** | **Val** | **Test** |
| **No. of screening exams** | 94363 | 11890 | 11593 | 14916 | 1860 | 1880 |
| No. of diagnostic exams | 25749 | 3207 | 3218 | 1777 | 203 | 243 |
| No. of women | 35265 | 4420 | 4392 | 11961 | 1471 | 1509 |
| Age | | | | | | |
| <40 | 2044 | 253 | 254 | 0 | 0 | 0 |
| 40-49 | 27873 | 3600 | 3333 | 135 | 17 | 12 |
| 50-59 | 41974 | 5372 | 5392 | 5740 | 683 | 729 |
| 60-69 | 31306 | 4035 | 3752 | 6925 | 875 | 884 |
| 70-79 | 14256 | 1595 | 1809 | 3663 | 457 | 468 |
| ≥80 | 2658 | 242 | 271 | 230 | 31 | 30 |
| Mean (std) | 57.4 (10.5) | 56.9 (10.2) | 57.3 (10.4) | 63.3 (7.3) | 63.5 (7.3) | 63.4 (7.2) |
| Median | 57.0 | 56.0 | 57.0 | 63.0 | 63.0 | 63.0 |
| Race | | | | | | |
|   Black | 34288 | 4300 | 3979 | 570 | 82 | 80 |
|   Asian or Pacific Islander | 2071 | 281 | 310 | 875 | 119 | 108 |
|   White | 73687 | 9150 | 9286 | 13495 | 1632 | 1694 |
|   Other | 10066 | 1366 | 1236 | 1753 | 230 | 251 |
| BI-RADS density | | | | | | |
|   A | 10971 | 1402 | 1282 | N/A | N/A | N/A |
|   B | 52418 | 6435 | 6486 | NA | N/A | N/A |
|   C | 32026 | 4146 | 4005 | N/A | N/A | N/A |
|   D | 3902 | 473 | 408 | N/A | N/A | N/A |
|   Unknown | 20795 | 2641 | 2630 | 16693 | 2063 | 2123 |

Table 1 shares the demographics of the 247,736 screening and 62,832 diagnostic exams from 93,450 women. Notably, OPTIMAM had a higher average age than WUSTL and a higher percentage of Asian or Pacific Islander women, while WUSTL had a higher percentage of black women. Finally, BI-RADS density is unknown for OPTIMAM.

## 3.2 Cancer Prevalence in Dataset Attributes

Table 2. Prevalence of cancer (exam counts) along stratifications of different attributes, including use of view markers, data source (WUSTL or OPTIMAM), exam type (screening or diagnostic), and scanner model (Hologic Selenia or Selenia Dimensions [Hologic Inc, Marlborough, MA]). Supplementary Table 1 shows the prevalence and counts for non-cancers. Dataset bias exists, as there are large differences in cancer prevalence across attribute values. [a]HS: Hologic Selenia. [b]SD: Selenia Dimensions (SD). [c]The exam counts are for screening and diagnostic exams captured on all scanners. [d]These are image counts instead of exam counts.

|  | WUSTL, % cancer prevalence (N) | | | OPTIMAM, % cancer prevalence (N) | | |
| --- | --- | --- | --- | --- | --- | --- |
|  | Train | Val | Test | Train | Val | Test |
| **Screening HS**[a] | 0.5 (143) | 0.3 (12) | 0.3 (14) | 28.0 (3937) | 27.7 (484) | 29.8 (534) |
| **Screening SD**[b] | 1.1 (707) | 1.2 (95) | 1.2 (88) | 57.4 (487) | 59.1 (65) | 64.4 (56) |
| **Screening All**[c] | 0.9 (850) | 0.9 (107) | 0.9 (102) | 29.7 (4424) | 29.5 (549) | 31.4 (590) |
| **Diagnostic All**[c] | 10.2 (2617) | 9.5 (306) | 10.2 (327) | 69.3 (1231) | 67.0 (136) | 74.1 (180) |
| **View Markers All**[d] | 1.2 (4504) | 1.2 (563) | 1.2 (546) | 20.4 (443) | 18.7 (50) | 18.3 (44) |
| **No View Markers All**[d] | 2.6 (3141) | 2.1 (321) | 2.7 (409) | 17.7 (10573) | 17.2 (1271) | 19.2 (1451) |

WUSTL had a natural prevalence of 0.9% while OPTIMAM had a cancer-enriched prevalence of 31.4% (Table 2). Both datasets saw a higher prevalence for diagnostic exams: 10.2% for WUSTL and 74.1% for OPTIMAM. Screening exams saw an increased prevalence from HS to SD: WUSTL, 0.3% to 1.2%, and OPTIMAM, 29.8% to 64.4%. This stratification revealed attributes with potentially exploitable dataset bias.

## 3.3 Shortcuts

### 3.3.1 View markers

View markers are burned-in annotations indicating the laterality and view of an image (Supplementary Figure 2). We observed a slight difference in the cancer prevalence for images with and without view markers for OPTIMAM (20.4% vs 17.7%) and a twofold difference for the WUSTL (1.2% vs 2.6%) (Table 1). We then trained a model to predict cancer using images with only the view markers (all breast tissue was removed). This model achieved an OPTIMAM AUC of 0.542 [95% CI 0.514, 0.572] and a WUSTL AUC of 0.691 [0.644, 0.736]. For WUSTL, the model learned the absence of view markers doubled the cancer prevalence and achieved a better than random performance, even without seeing any breast tissue. To mitigate this shortcut, we removed any markings outside of the breast during training and adopted this strategy for all subsequent models.

### 3.3.2 Dataset

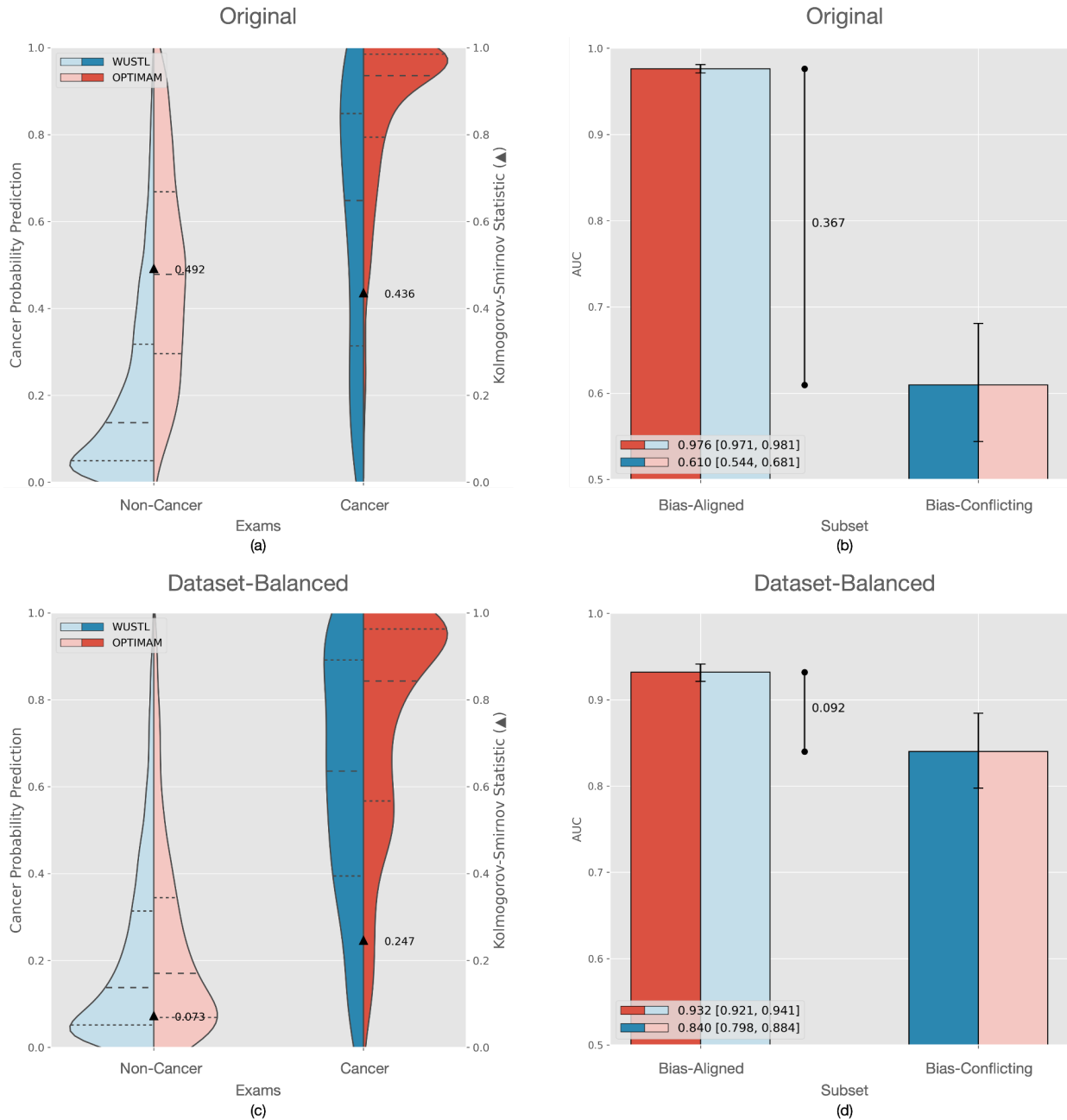

Figure 2: Figures (a,b) and (c,d) are for the original and dataset-balanced models, respectively. Figures (a,c) are violin plots of the model prediction distributions on WUSTL and OPTIMAM, stratified by non-cancer and cancer exams. The dashed lines represent the quartiles of the exam scores. The triangle indicates the value of the KS statistic, representing the maximum distance between the cumulative distribution functions of WUSTL and OPTIMAM. As seen by the large KS statistic, OPTIMAM exams have higher scores than WUSTL exams, but this difference is mitigated when dataset-balancing is applied, as seen in the decrease of the KS statistic. Figures (b,d) show the AUCs [95% CI] for the constructed pathological, complementary subsets. The left bar is the bias-aligned subset of WUSTL non-cancer and OPTIMAM cancer exams. The right bar is the bias-conflicting subset of WUSTL cancer and OPTIMAM non-cancer exams. The difference between the bias-aligned and bias-conflicting AUCs are in between the bars. The difference between these two subsets is reduced when using dataset-balancing, suggesting a decrease in the model bias.

The original model achieved a 0.945 AUC on the combined WUSTL+OPTIMAM dataset, paradoxically higher than either individual AUC: 0.838 (WUSTL) and 0.892 (OPTIMAM) (Table 3). Initial investigation confirmed a significantly higher cancer prevalence in OPTIMAM (31.4%) than WUSTL (0.9%) (Table 2). To decorrelate this dataset bias, we equally sampled cancers and non-cancers from WUSTL and OPTIMAM during training. This dataset-balanced model increased AUC to 0.861 (WUSTL) and 0.918 (OPTIMAM), while the combined AUC decreased to 0.920 (Table 3). While the combined AUC was still higher than that of the individual datasets, the gap was significantly reduced.

On average, the cancer predictions were higher in OPTIMAM for both cancer and non-cancer exams (Figure 2a), indicating the presence of model bias. Furthermore, a DL model trained to predict the dataset source from the FFDM images achieved a 0.954 AUC (Table 4), suggesting that the model was able to identify and exploit differences in the appearance of the images in the two datasets. Dataset-balancing reduced this model bias: the KS statistic, the maximum distance between two distributions, decreased between WUSTL and OPTIMAM (Figure 2c). The gap between the bias-aligned (OPTIMAM cancers and WUSTL non-cancers) and the bias-conflicting subsets (OPTIMAM non-cancers and WUSTL cancers) also reduced from 0.367 to 0.092 (Figure 2b, 2d). A LR model predicting dataset source from the malignancy model's feature vector dropped from 0.872 to 0.692 AUC, suggesting that dataset-balancing attenuates the utilization of dataset source in the malignancy prediction (Table 4). We adopted this dataset-balanced strategy for the remaining experiments.

### 3.3.3 Exam type

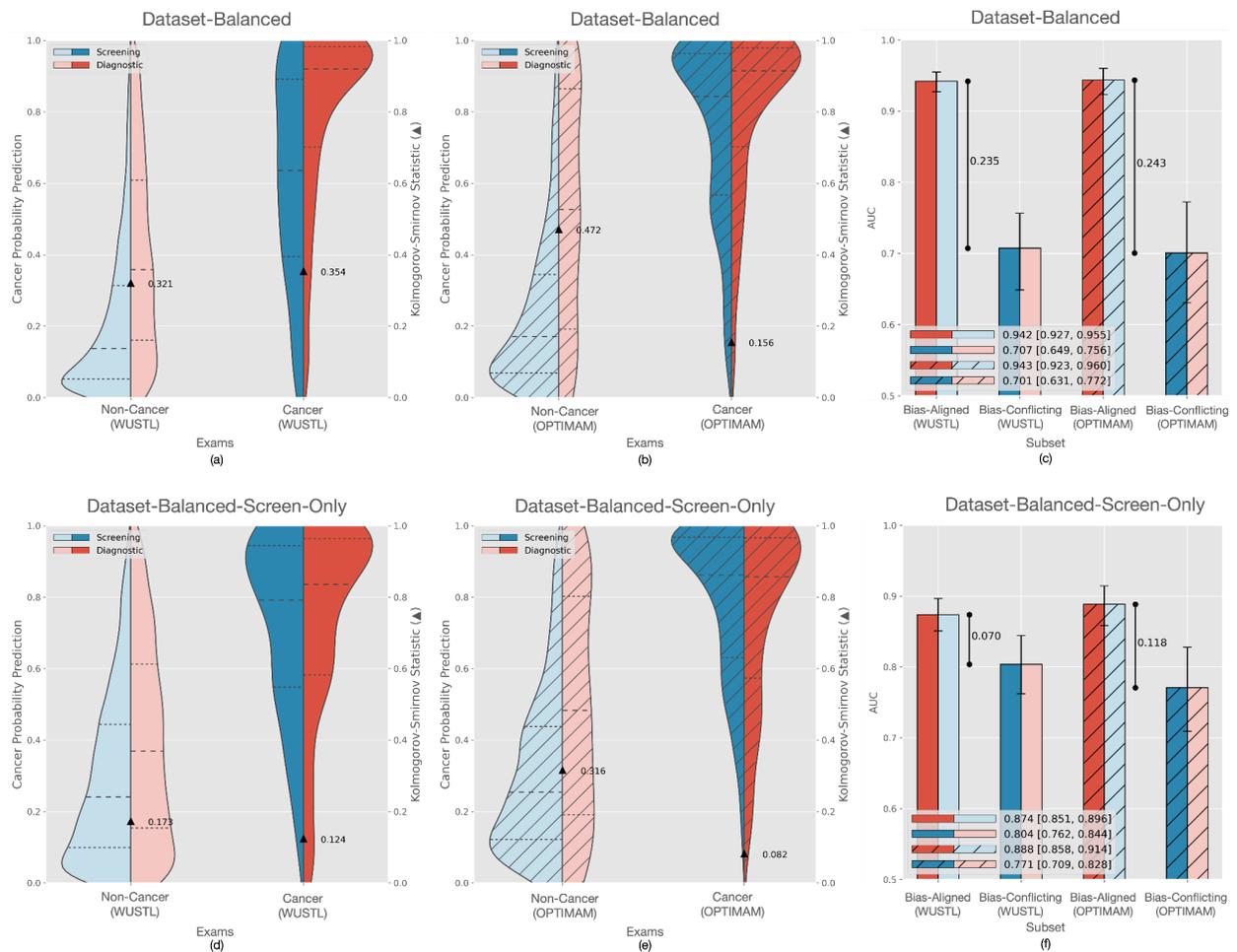

Figure 3: Figures (a,b,c) and (d,e,f) are the exam type shortcuts for the dataset-balanced and dataset-balanced-screen-only models, respectively. Figures (a,b) and (d,e) are violin plots over the WUSTL and OPTIMAM datasets, respectively. The dashed lines represent the quartiles over the exam scores. The KS statistic is presented as a triangle in the violins. Except for OPTIMAM cancers, diagnostic exams overall have a higher score than screening exams. This effect is reduced by only using screening exams in training. Figures (c,f) give the AUCs [95% CI] of the biased-aligned (screening non-cancers and diagnostic cancers) vs the bias-conflicting (screening cancers and diagnostic non-cancers) for WUSTL and OPTIMAM. The difference between the bias-aligned and bias-conflicting AUCs are provided between the bars. The gap between the bias-aligned and bias-conflicting subsets is reduced when we train only on screening exams, suggesting a reduction in model bias.

Use of diagnostic exams increases the number of cancer examples during training; however, WUSTL evaluation unveiled an AUC paradox with 0.903, 0.861, and 0.862 AUCs on the combined, screening, and diagnostic sets, respectively (Table 3). In WUSTL, diagnostic exams had ten times the cancer prevalence: 10.2% vs 0.9% (Table 2). In OPTIMAM, the performance on the combined dataset (0.915) was not higher than both the screening (0.918) and diagnostic (0.764) datasets (Table 3), likely due to having only 243 diagnostic exams (Table 1). Consequently, training a DL model to predict exam type achieved a 0.798 AUC. To mitigate this shortcut, we train only using screening exams to improve the screening AUC to 0.882 and 0.919 on WUSTL and OPTIMAM, respectively (Table 3).

In both datasets, training solely on screening exams reduced the KS statistic between diagnostic and screening exams (Figures 3a, 3b, 3d, 3e) and attenuated the differences between the bias-aligned and bias-conflicting subsets (Figures 3c, 3f). While the LR model achieved 0.723 AUC at predicting exam type, the shortcut mitigation reduced the LR AUC to 0.632. Our remaining experiments utilized this dataset-balanced-screen-only strategy.

### 3.3.4 Scanner Models

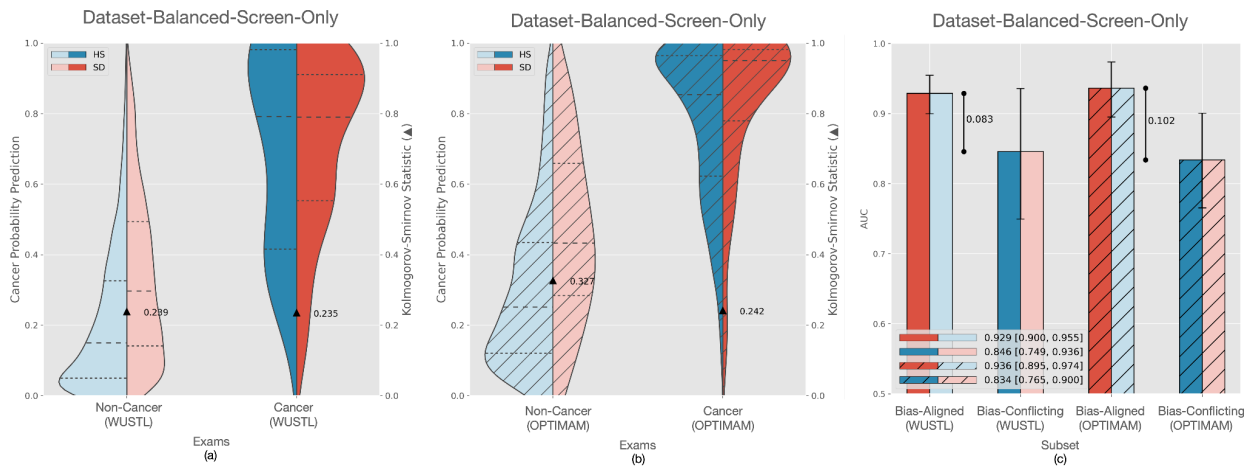

Figure 4. The predictions from the dataset-balanced-screen-only model. Figures (a) and (b) are the prediction violin plots for WUSTL and OPTIMAM stratified on the scanner model. The KS statistic reveals that the model has learned high model bias for SD scanners in general. Figure (c) shows the AUCs [95% CI] for the constructed bias-aligned subset of SD cancers and HS non-cancers and the complementary bias-conflicting subset. The difference between the bias-aligned and bias-conflicting AUCs are provided between the bars. Note how the model exhibits shortcut biased behavior for both datasets.

Although SD is the newer scanner (see Supplementary Figure 3 for example images), the dataset-balanced-screen-only model performed better on HS scanners, likely due to the higher number of HS cancers during training. For the HS, SD, and combined datasets, the model achieved 0.925, 0.860, and 0.882 AUC, respectively, for WUSTL and 0.919, 0.882, and 0.919, respectively, for OPTIMAM (Table 3). While no AUC paradox was observed for this shortcut, both WUSTL (1.2% vs 0.3%) and OPTIMAM (64.4% vs 29.8%) manifested a dataset bias favoring SD exams.

In general, SD exams had higher scores than HS exams. However, for WUSTL cancer exams, the distinction was less clear, as HS only had 14 cancer exams (Figures 4a, 4b). The bias-aligned subset had a slightly higher performance than the bias-conflicting subset, with a 0.083 difference for WUSTL and 0.102 for OPTIMAM (Figure 4c). The image DL and LR models achieved a 0.978 and 0.680 AUC, implying the use of scanner model features in malignancy prediction.

Table 3. Summary table of the stratification of the malignancy task performance for the original, dataset-balanced, and dataset-balanced-screen-only models over the scanner stratified, screening, diagnostic, and screening+diagnostic datasets. For both WUSTL and OPTIMAM, the dataset-balanced screen-only model performs the best on Screening HS, Screening SD, and Screening All, except for OPTIMAM screening SD. The original model appears to perform the best on WUSTL+OPTIMAM Screening All due to the AUC paradox. The dataset-balanced model performed the best on diagnostic and the shortcut-prone screening+diagnostic; however, our task is focused on screening exams. Bolded entries are the best (highest). Metric is AUC [95% confidence interval]. HS: Hologic Selenia scanner; SD: Selenia Dimensions scanner.

|  | **Original** | **Dataset-balanced** | **Dataset-balanced screen-only** |
|---|---|---|---|
| **Screening HS** |  |  |  |
| WUSTL | 0.913 [0.816, 0.982] | 0.886 [0.766, 0.979] | **0.925 [0.867, 0.978]** |
| OPTIMAM | 0.890 [0.872, 0.906] | 0.916 [0.900, 0.929] | **0.919 [0.905, 0.932]** |
| **Screening SD** |  |  |  |
| WUSTL | 0.806 [0.749, 0.857] | 0.843 [0.798, 0.886] | **0.860 [0.819, 0.899]** |
| OPTIMAM | 0.895 [0.822, 0.957] | **0.931 [0.877, 0.978]** | 0.882 [0.810, 0.948] |
| **Screening All** |  |  |  |
| WUSTL + OPTIMAM | **0.945 [0.936, 0.954]** | 0.920 [0.909, 0.930] | 0.905 [0.893, 0.917] |
| WUSTL | 0.838 [0.792, 0.878] | 0.861 [0.818, 0.898] | **0.882 [0.844, 0.912]** |
| OPTIMAM | 0.892 [0.875, 0.906] | 0.918 [0.905, 0.931] | **0.919 [0.905, 0.930]** |
| **Diagnostic All** |  |  |  |
| WUSTL | 0.855 [0.828, 0.881] | **0.862 [0.837, 0.884]** | 0.809 [0.780, 0.838] |
| OPTIMAM | 0.745 [0.671, 0.809] | **0.764 [0.694, 0.832]** | 0.743 [0.670, 0.813] |
| **Screening + Diagnostic All** |  |  |  |
| WUSTL | 0.896 [0.879, 0.912] | **0.903 [0.886, 0.919]** | 0.862 [0.841, 0.881] |
| OPTIMAM | 0.891 [0.873, 0.907] | **0.915 [0.902, 0.926]** | 0.905 [0.893, 0.917] |

Table 4: Summary table of image-level AUC performance and [95% confidence intervals] of predicting the following attributes: the dataset source (WUSTL vs OPTIMAM), the exam type (screening vs diagnostic), and the scanner model (HS vs SD). The DL model (first row) was trained from scratch on the images to predict the attribute. The remaining models refer to logistic regression models trained on the pre-logit feature vector from that malignancy model. The features produced by the shortcut mitigated models predict attributes with lower AUCs, i.e. less fidelity indicating a successful reduction in shortcut exploitation. Bolded entries are the best (lowest). Averaging over all three attributes, the dataset-balanced-screen-only logits had the lowest vaerage AUC, but it did not always have the lowest AUC over individual attributes, revealing the complexity of completely removing shortcut attributes from training.

| Model | Attribute: Dataset | Attribute: Exam Type | Attribute: Scanner Model |
|---|---|---|---|
| DL Model | 0.954 [0.951, 0.956] | 0.798 [0.793, 0.803] | 0.978 [0.977, 0.979] |
| Original logits | 0.872 [0.868, 0.875] | 0.738 [0.733, 0.744] | **0.669 [0.665, 0.673]** |
| Dataset-balanced logits | **0.692 [0.686, 0.698]** | 0.727 [0.722, 0.733] | 0.723 [0.719, 0.727] |
| Dataset-balanced-screen-only logits | 0.726 [0.721, 0.731] | **0.642 [0.636, 0.648]** | 0.680 [0.676, 0.685] |

## 3.4 AUC Paradox

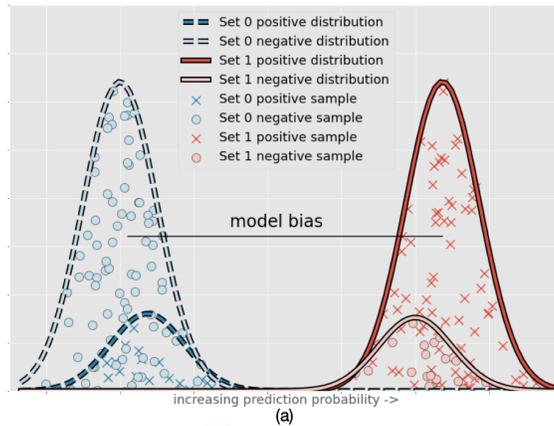

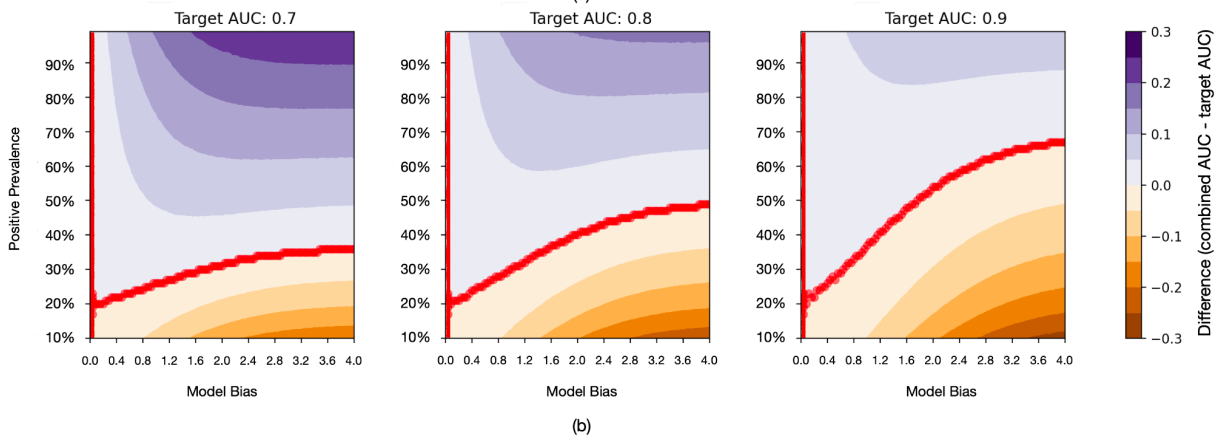

Figure 5: (a) An example of the binormal distributions used to generate samples for the two datasets with a 0.7 target AUC. Set 0 has a positive prevalence of 20% and no model bias, and Set 1 has an increased positive prevalence of 80% and a strong model bias of 8 that pushed the predicted probability (x-axis) to the right, leading to a 0.85 combined AUC. (b) Plots of the paradoxical difference in combined AUC and target AUC when varying target AUC, model bias, and positive prevalence (dataset bias) of Set 1 for target AUCs of 0.7, 0.8, and 0.9  Each point is the average of 100 simulations where both data sets are resampled (both dataset sizes are sampled randomly from 10,000 to 40,000). The unaugmented Set 0 has 20% positive prevalence and no model bias added. Set 1 had a positive prevalence from 90 linearly spaced values ranging from 10% to 99% and a model bias from 101 linearly spaced values from 0 to 4. The red points are the zero crossings, where the combined AUC was equal to the target AUC.

To understand the AUC paradox, we simulated two datasets with the binormal model[26] (Figure 5a), altering the positive prevalence and model bias of Set 1. The combined AUC appeared inflated when both the prevalence and the bias increased but deflated when the prevalence decreased and the bias increased (Figure 5b). Increasing the bias also increased the prevalence needed to reach the zero crossings (red dots), where the combined AUC was equal to the target AUC. Near these crossings, holding prevalence constant, the bias and AUC difference were negatively correlated; holding bias constant, the prevalence and AUC difference were positively correlated.

AUC is the probability that a random positive score is higher than a random negative score.[30–32] When the cancer prevalence is higher in Set 1, a model may learn to predict higher scores for this dataset (model bias). Intuitively, concatenating these datasets inflated the AUC because most positives came from the biased Set 1 and most negatives came from the unbiased Set 0. This effect could also lead to a deflated AUC if for some reason there was a negative bias for the dataset with higher cancer prevalence. To attenuate this inflation or deflation, one can evaluate on (1) stratified datasets or (2) datasets with the same positive prevalence when the model bias is low (e.g., <0.4) (Supplementary Section 4 and 5).

# 4 Discussion

This work examines how different cancer prevalences within datasets lead to shortcuts and obscures performance of DL models in screening mammography. We present techniques to identify spurious correlations between different dataset attributes and cancer prevalence. We demonstrated that four shortcuts, view markers, dataset source, exam type, and scanner model, were exploited during training and mitigated three shortcuts by removing view markers, sampling equally from datasets, and training on screening exams only. We also investigated how combining datasets can alter performance via the AUC paradox. Understanding shortcuts and dataset compositions are crucial for obtaining an accurate estimate of a model's performance, especially when these models are trained using disease-enriched datasets. These potential pitfalls in training and evaluation are vital to understand as AI systems move beyond the laboratory into real world clinical use.

## 4.1 Shortcuts

Differences in cancer prevalence along attribute values reveals dataset bias. The output probability distributions, bias-aligned vs bias-conflicting subsets, and the attribute value prediction methods reveal the corresponding model bias. These methods can identify shortcuts even when there is no difference in stratified AUCs, an important aspect since model bias can shift the entire distribution and therefore require different thresholds for different attributes. These shortcuts were ameliorated by either removing the attribute or balancing the cancer prevalence between attribute values during training. Identifying and mitigating shortcuts are crucial for real-world generalization, as shortcuts may not transfer to other sites, deteriorating performance.

## 4.2 AUC Paradox

Our analysis revealed that combining datasets with dataset and model biases can spuriously inflate or deflate AUCs. These findings explain our shortcut observations, as there were clear dataset and model biases for OPTIMAM and diagnostic exams. Although we did observe biases for scanner models, we did not observe the AUC paradox, possibly due to the limited number of HS exams in WUSTL and SD exams in OPTIMAM. Even for attributes where differences in cancer prevalence are clinically representative, such as screening and diagnostic exams, isolated estimates of performance are preferred over aggregate statistics, as they are less likely to hide overestimates in performance arising from shortcut learning.

At the highest level, this finding has ramifications on how to evaluate and regulate AI devices to ensure safety and efficacy to patients. We observed that the combined and target AUC are equal across a range of prevalences only when bias is exactly 0. Since this is unlikely for any real AI model, we consider the low bias case, which was trained by removing biased attributes and balancing cancer prevalence during training. In this simulation, for a small model bias (e.g., <0.4), the combined AUC is the same as target AUC when the prevalence is similar between sets (e.g., 20%). For larger model bias, when the prevalence is similar between sets, combined AUC is conservative and underestimates target AUC. Since pivotal trials may rely on a combined AUC, it may therefore be wise to consider balancing prevalence from the sites that make up the combined data.

Recognizing this paradox is critically important for radiologists when evaluating AI products. It can mean the overall performance presented is falsely higher than the reported performance at individual sites. Thus, when evaluating reported measures of AUC for an AI system, be cautious of results reported over aggregate data sets from disparate sources. If the individual sources vary in cancer prevalence, then the reported AUC may be misleading when compared to all sources individually.

### 4.3 Limitations

The lack of data in WUSTL was a key limitation: the screening test set had only 102 cancers, including 14 HS malignant exams. The small number of WUSTL HS exams hindered comparisons of performance for the scanner model and prevented investigation of scanner model balancing to mitigate the shortcut. Also, the training and testing datasets all came from the same sites. Additionally, our AUC paradox simulation assumes a normal distribution of scores, which may not accurately represent the model score distribution on real data. Finally, this work focused on shortcuts for attributes labeled with discrete classes (e.g. HS vs SD). In reality, these models may find shortcuts on hidden attributes or continuous attributes. Future studies will investigate debiasing methods not requiring pre-specification of the attributes of interest and propose shortcut mitigations without removing crucial cancer cases like diagnostic exams.

### 4.4 Conclusion

We identified spurious shortcuts taken by deep learning models during training and evaluation issues that artificially inflate the performance and hinder model generalizability. Neglecting to assess shortcuts during any phase of (1) selecting the best model, (2) selecting an operating threshold for the decision system, (3) conducting retrospective clinical studies, or (4) evaluating performance in clinical settings can be dangerous. The methods and techniques presented in this work are a broadly applicable framework that should be considered during development of any AI system for radiology to minimize risks and maximize benefits to the future patients and practices depending on these models.

# Acknowledgements

We thank Andrea R. Gwosdow, Ph.D. of Gwosdow Associates Science Consultants, LLC for assistance in preparing this manuscript on behalf of Whiterabbit AI, Inc.

# Financial Disclosure

This work was funded by Whiterabbit AI, Inc.

# Supplementary Information - Problems and shortcuts in deep learning for screening mammography

## Supplementary Section 1: Data

### S.1.1 Non-cancer data

We explored mammography exams from two institutions: Washington University in St. Louis (WUSTL) from the US and the National Health Service OPTIMAM database[1] (OPTIMAM) from the UK. Supplementary Table 1 shares the prevalence and exam count of non-cancer exams for both of these datasets stratified across the scanner models and exam types, revealing clear differences in prevalence.

Supplementary Table 1: Prevalence of non-cancer (exam counts) for two datasets: WUSTL and OPTIMAM.

|  | WUSTL, % non-cancer prevalence (N) | | | OPTIMAM, % non-cancer prevalence (N) | | |
|---|---|---|---|---|---|---|
|  | Train | Val | Test | Train | Val | Test |
| **Screening HS**[a] | 99.5 (31612) | 99.7 (3936) | 99.7 (3994) | 72.0 (10129) | 72.3 (1266) | 70.2 (1259) |
| **Screening SD**[b] | 98.9 (61899) | 98.8 (7847) | 98.8 (7497) | 42.6 (361) | 40.9 (45) | 35.6 (31) |
| **Screening All**[c] | 99.1 (93513) | 99.1 (11783) | 99.1 (11491) | 70.3 (10492) | 70.5 (1311) | 68.6 (1290) |
| **Diagnostic All**[c] | 89.8 (23132) | 90.5 (2901) | 89.8 (2891) | 30.7 (546) | 33.0 (67) | 25.9 (63) |
| **View Markers**[d] | 98.8 (372481) | 98.8 (47290) | 98.8 (45934) | 79.6 (1730) | 81.3 (217) | 81.7 (197) |
| **No View Markers**[d] | 97.4 (118548) | 97.9 (15128) | 97.3 (14773) | 82.3 (49056) | 82.8 (6113) | 80.8 (6115) |

[a]HS: Hologic Selenia. [b]SD: Selenia Dimensions (SD). [c]The exam counts are for screening and diagnostic exams captured on all scanners. [d]These are image counts instead of exam counts. Note: HS and SD do not add up to All, as there are exams with multiple scanner models.

## S.1.2 View markers

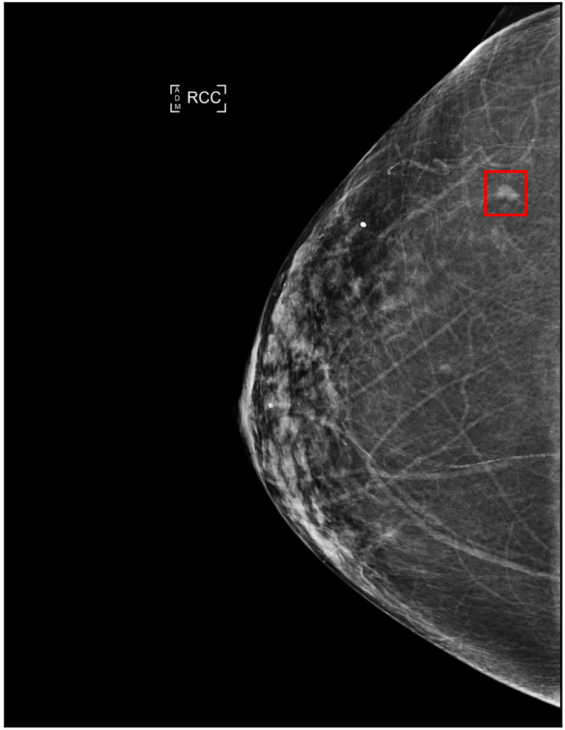 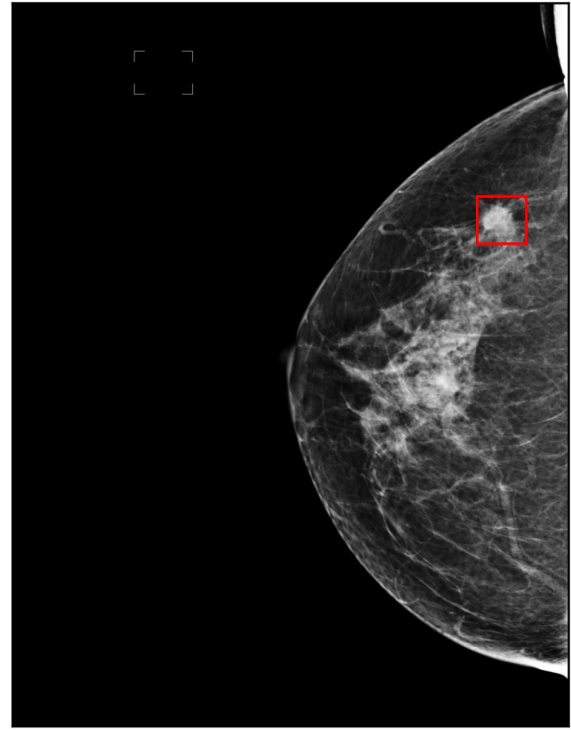

(a) WUSTL  (b) OPTIMAM

Supplementary Figure 1: View markers on the top left of the image on WUSTL and OPTIMAM, respectively. The malignant finding is located within the red bounding box.

Supplementary Figure 1 shows an example of view markers on WUSTL and OPTIMAM. WUSTL view markers are brighter with the image view and text, while OPTIMAM view markers are dimmer with no text. These clear differences can be exploited by a model to learn the difference between WUSTL and OPTIMAM.

## S.1.3 Scanner models

Supplementary Figure 2 shows an exam from HS and SD from WUSTL and OPTIMAM. While breast tissues change over time and positions may be different, the HS images (top row) and their corresponding SD images (bottom row) also differ in subtle image features, like contrast and brightness. These subtle features are easily distinguishable for a deep learning (DL) model, and when certain attribute values are correlated with cancer, it becomes easier to distinguish these attribute values instead of learning actual cancer pathology.

# (a) WUSTL

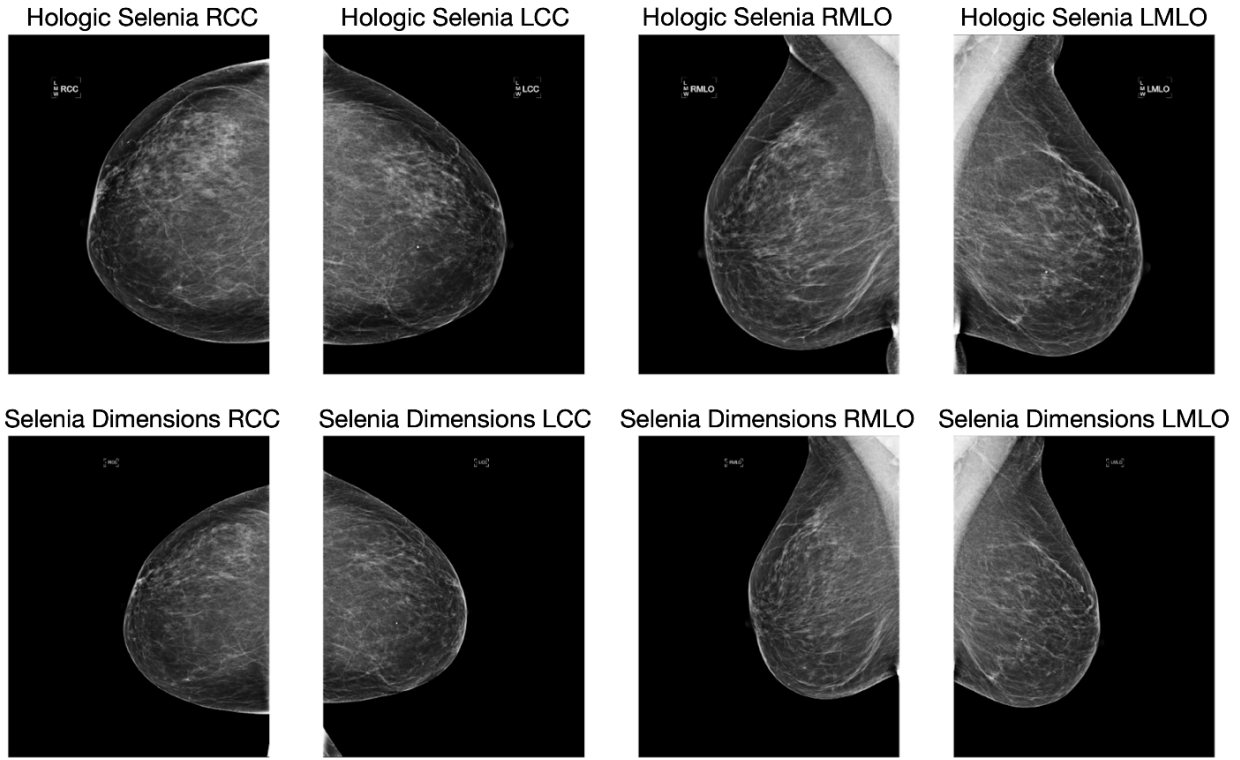

# (b) OPTIMAM

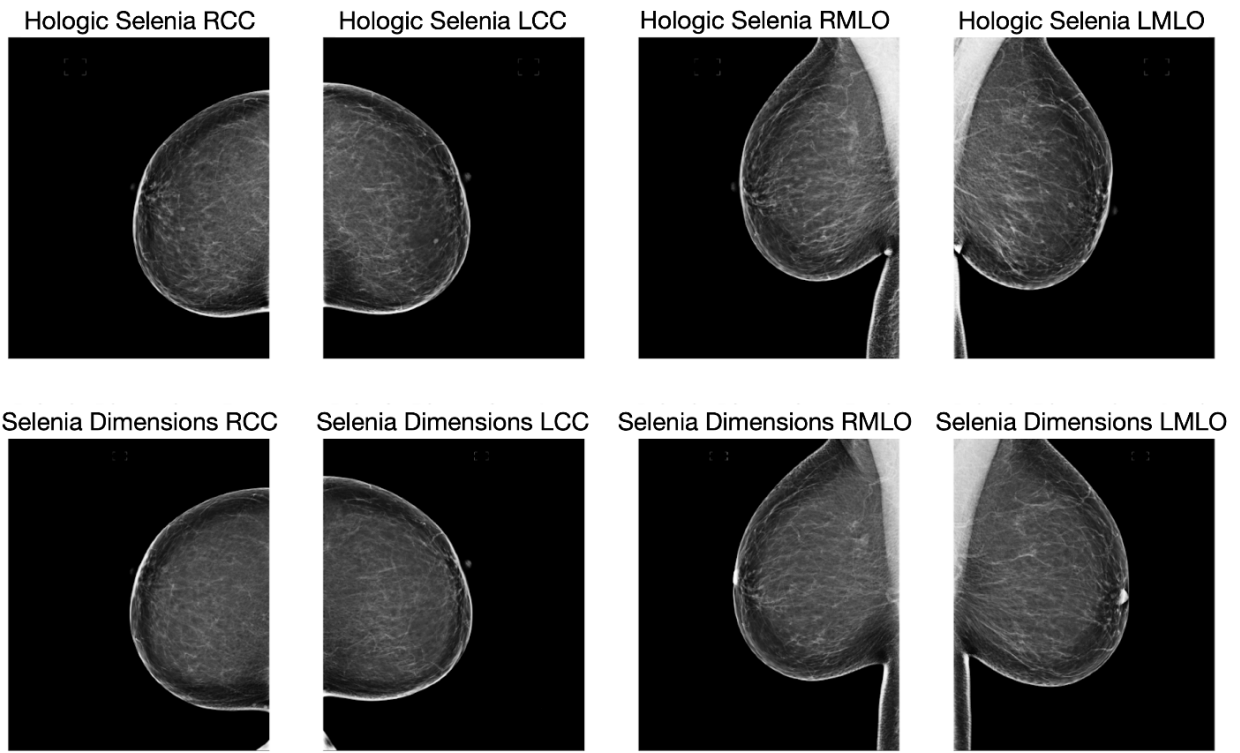

Supplementary figure 2: Example images from the Hologic Selenia (top row) and Selenia Dimensions (bottom rows) for the (a) WUSTL and (b) OPTIMAM datasets.

## Supplementary Section 2: Malignancy Model Training

Our malignancy model was an Efficient-B0 model initialized using ImageNet[2] weights. We trained this model in two stages: patch training and whole-image training. For the patch model, the model is trained such that each mini batch of 32 patches contains a 1:1 ratio of cancer to non-cancer samples. For cancers, we create a patch around the center of the radiologist-provided annotation indicating the location of the malignant finding. For non-cancers, we create a patch from a randomly sampled region in the image of a non-cancer breast. Patches are clipped from 0 to 4095 and then normalized from 0 to 1. During training, patches are: (1) randomly flipped both vertically and horizontally with 50% probability, (2) translated at most 50 pixels in a random direction, (3) randomly rotated between -45 and 45 degrees; and (4) have Gaussian noise with a standard deviation of 5e-3 added. If the previous operations would result in a patch extending beyond the extent of the image, the patch is translated such that the entire area is within the image. Training is performed with the Adam[3] optimizer with an initial learning rate of 1e-3 and weight decay of 1e-4. The learning rate is reduced by a factor of 0.967 every epoch, which consists of 25,000 patches. We select the epoch with the highest average validation AUC of WUSTL and OPTIMAM.

The weights of the patch model are used to initialize the whole-image model. This is a common approach to use for image-level malignancy classification. The images are resized from their original resolutions to 1664x1280 pixels through bilinear interpolation. This represents roughly half the native resolution. The remaining training pipeline mirrors that of the patch model. The only difference is that the image undergoes a different translation distribution (-10% to +10% of the image dimensions), a different rotation distribution (-10 to +10 degrees), and is additionally further scaled to between 90% and 110% of the original dimensions prior to the resizing to 1664x1280 pixels. As with the aforementioned patch model, we select the epoch with the highest average validation AUC of WUSTL and OPTIMAM.

This architecture produces predictions for individual images. However, we want to evaluate the performance at the exam level. While training directly on exams can be done, we opt for this simplified setup, as there are many potential approaches for the design of a model that operates directly on exams. Instead, we train a deep learning model to classify an image, and we use intuitive aggregations to calculate an exam-level probability. First, an exam may consist of multiple images of a breast corresponding to different views. The predictions for each image of a breast are averaged together to create a breast probability, as they may contain related information. Then, the maximum breast probability of the two breasts becomes the exam-level probability. This is done as cancer is typically found in only one breast. This overall setup was held constant for all experiments to ensure that conclusions are not caused by the alteration of hyperparameters or configurations.

## Supplementary Section 3: Attribute Prediction Models

The DL models were trained on images to predict the binary value for a given attribute: WUSTL vs OPTIMAM, HS vs SD, and screening vs diagnostic. They used the same pipeline as the malignancy models with two differences. Firstly, the pre-trained ImageNet weights were used for initialization instead of pre-training on patches. Secondly, each mini batch during training was set to a 1:1 ratio based on the attribute values.

The logistic regression models trained to predict the attribute value require the learned representations of a malignancy model. We use the feature vector resulting from the last global pooling operation in the EfficientNet-B0 architecture and before the final logit layer. This produces a 1280 dimensional vector for a given input image. These vectors were generated for the entirety of the val and test sets for both WUSTL and OPTIMAM. The logistic regression models were trained for 500 iterations on the val sets' vectors with a L2 regularization penalty and then evaluated on the test sets' vectors to produce the AUC values reported.

# Supplementary Section 4: AUC with the Binormal Model

## S.4.1 Data Simulation with the Binormal ROC Model

**Proposition 1.** *Let $X_0 \sim \mathcal{N}(\mu_0, \sigma_0^2)$ and $X_1 \sim \mathcal{N}(\mu_1, \sigma_1^2)$ be independent random variables, and let $\text{AUC} := P(X_1 > X_0)$. Then, $\text{AUC} = \Phi\left(\frac{\mu_1 - \mu_0}{\sqrt{\sigma_1^2 + \sigma_0^2}}\right)$*

*Proof.* Let $D := X_1 - X_0$. We have $D \sim \mathcal{N}(\mu_1 - \mu_0, \sigma_1^2 + \sigma_0^2)$ since $X_0$ and $X_1$ are jointly normal. Thus,

$$\text{AUC} = P(X_1 > X_0) = P(D > 0) = P\left(Z > \frac{-\mu_1 + \mu_0}{\sqrt{\sigma_1^2 + \sigma_0^2}}\right) = 1 - \Phi\left(\frac{-\mu_1 + \mu_0}{\sqrt{\sigma_1^2 + \sigma_0^2}}\right),$$

where $Z \sim \mathcal{N}(0, 1)$ and $\Phi$ denotes the standard normal CDF. Since $\Phi$ is symmetric, we have $\text{AUC} = \Phi\left(\frac{\mu_1 - \mu_0}{\sqrt{\sigma_1^2 + \sigma_0^2}}\right)$, as desired. □

Supplementary Figure 3: Proposition 1 - AUC of the binormal ROC model.

In this study, we generate simulated data according to the conventional binormal ROC model following Pan et al. (1997)[4]. For the unbiased Set 0, the scores for the negative class are sampled from $\mathcal{N}(0, 1)$ whereas the scores for the positive class are sampled from $\mathcal{N}(a, 1)$, with $a \in \mathbb{R}$ a fixed parameter. Applying Proposition 1 (Supplementary Figure 3), there is an analytic relationship between AUC and parameter $a$ :

$$\text{AUC} = \Phi\left(\frac{a}{\sqrt{2}}\right) \iff a = \sqrt{2}\Phi^{-1}(\text{AUC})$$

where $\Phi$ is the cumulative distribution function (CDF) of the standard normal distribution, and $\Phi^{-1}$ is the inverse of the CDF — the percent point function. Thus, given a target AUC, this relationship allows us to solve for $a$ to achieve a simulated dataset with that AUC.

For the biased Set 1, we add a model bias $m \in \mathbb{R}_0^+$. Specifically, scores for the negative class are sampled from $\mathcal{N}(m, 1)$, whereas scores for the positive class are sampled from $\mathcal{N}(a + m, 1)$. Again, applying Proposition 1 (Supplementary Figure 3), the AUC for Set 1 is the same with the AUC for Set 0, i.e. $\Phi\left(\frac{a}{\sqrt{2}}\right)$.

## S.4.2 Difference between Combined AUC and Target AUC

We provide a probabilistic framework to support the trends seen in the simulated results (Section 3.4, main paper). Recall that AUC represents the probability that a randomly chosen positive sample has a greater score than a randomly chosen negative sample.[5–7] We will examine the difference between the AUC of the combined datasets $AUC_C$ and the target AUC $AUC_T$ of the individual datasets:

**Theorem 1.** *Let $S_0 \sim \text{Bernoulli}(p_0)$ and $S_1 \sim \text{Bernoulli}(p_1)$ specify the sets from which negative and positive examples are sampled, respectively. Let $\text{AUC}_T$ denote the fixed target AUC and $a = \sqrt{2}\Phi^{-1}(\text{AUC}_T) \in \mathbb{R}$.*

*Let $X_0$ and $X_1$ represent the model scores on the negative and positive example, respectively. For Set 0 without model bias, $X_0$ and $X_1$ satisfy*

$$(X_0|S_0 = 0) \sim \mathcal{N}(0,1), \ (X_1|S_1 = 0) \sim \mathcal{N}(a,1),$$

*while, for Set 1 with model bias $m \in \mathbb{R}_0^+$,*

$$(X_0|S_0 = 1) \sim \mathcal{N}(m,1), \ (X_1|S_1 = 1) \sim \mathcal{N}(a+m,1).$$

*Let the combined $\text{AUC}_C \coloneqq P(X_1 > X_0)$. Then, the difference between combined and target AUC is*

$$\text{AUC}_C - \text{AUC}_T = \left[\Phi\left(\frac{a+m}{\sqrt{2}}\right) - \Phi\left(\frac{a}{\sqrt{2}}\right)\right](1-p_0)p_1 + \left[\Phi\left(\frac{a-m}{\sqrt{2}}\right) - \Phi\left(\frac{a}{\sqrt{2}}\right)\right]p_0(1-p_1).$$

*Proof.* Marginalizing $\text{AUC}_C = P(X_1 > X_0)$ over $S_0$ and $S_1$, we have

$$\text{AUC}_C = \sum_{i=0}^{1}\sum_{j=0}^{1} P(X_1 > X_0, S_0 = i, S_1 = j) \tag{1}$$

$$= \sum_{i=0}^{1}\sum_{j=0}^{1} P(X_1 > X_0 | S_0 = i, S_1 = j)P(S_0 = i, S_1 = j) \tag{2}$$

$$= \sum_{i=0}^{1}\sum_{j=0}^{1} \text{AUC}_{S_0=i,S_1=j}P(S_0 = i, S_1 = j). \tag{3}$$

Applying Prop. 1 to $(X_0|S_0 = i), (X_1|S_1 = i)$, we can easily see that when both $X_0$ and $X_1$ are both sampled from the same Set $i$, we achieve the target $\text{AUC}_{S_0=i,S_1=j} = \text{AUC}_T$. Thus, for the diagonal term of Eq. (3), we have

$$\sum_{i=j=0}^{1} \text{AUC}_{S_0=i,S_1=j}P(S_0 = i, S_1 = j) = \text{AUC}_T\left[P(S_0 = S_1 = 0) + P(S_0 = S_1 = 1)\right] \tag{4}$$

$$= \text{AUC}_T\left[1 - P(S_0 = 0, S_1 = 1) - P(S_0 = 1, S_1 = 0)\right]. \tag{5}$$

Substituting Eq. (5) to Eq. (3) and rearranging, we have

$$\text{AUC}_C - \text{AUC}_T = \text{AUC}_{S_0=0,S_1=1}P(S_0 = 0, S_1 = 1) + \text{AUC}_{S_0=1,S_1=0}P(S_0 = 1, S_1 = 0)$$
$$= \text{AUC}_{S_0=0,S_1=1}(1-p_0)p_1 + \text{AUC}_{S_0=1,S_1=0}p_0(1-p_1),$$

where the second equality is because $S_0$ and $S_1$ are independent Bernouilli variables. Finally, applying Prop. 1 to the two pairs $(X_0|S_0 = 0), (X_1|S_1 = 1)$, and $(X_0|S_0 = 1), (X_1|S_1 = 0)$, we obtain

$$\text{AUC}_C - \text{AUC}_T = \left[\Phi\left(\frac{a+m}{\sqrt{2}}\right) - \Phi\left(\frac{a}{\sqrt{2}}\right)\right](1-p_0)p_1 + \left[\Phi\left(\frac{a-m}{\sqrt{2}}\right) - \Phi\left(\frac{a}{\sqrt{2}}\right)\right]p_0(1-p_1),$$

as desired. $\square$

Supplementary Figure 4: Theorem 1 - The difference between the combined and target AUC.

Let us analyze the implications of Theorem 1. When $m = 0$, i.e. without model bias, we observe that the combined AUC is equal to the target AUC. When $m > 0$, since the CDF function $\Phi$ is a monotonically non-decreasing function, we observed the first term $[\Phi(\frac{a+m}{\sqrt{2}}) - \Phi(\frac{a}{\sqrt{2}})] \geq 0$ and second term $[\Phi(\frac{a-m}{\sqrt{2}}) - \Phi(\frac{a}{\sqrt{2}})] \leq 0$. Therefore, to have a combined AUC higher than the target AUC, the weighted first term must be greater than the weighted second term. The first term is weighted by $(1-p_0)p_1$, i.e. the probability we select a negative sample from Set 0 and a positive sample from Set 1. Intuitively, these are the bias-aligned examples: they boost AUC due to the positive first term. The second term is weighted with $p_0(1-p_1)$, i.e. the probability we select a negative sample from Set 1 and a positive sample from Set 0. Intuitively, these are the bias-conflicting samples: they pull AUC down due to the negative second term. As the target AUC $a$ or model bias $m$ increases, the CDF of the first term asymptotically approaches 1, requiring a higher weight than the second term to achieve a positive difference between combined AUC and target AUC. When $m < 0$, these terms are flipped. We investigate this theorem through a simulation in Supplementary Section 5.

# Supplementary Section 5: AUC Paradox Simulation

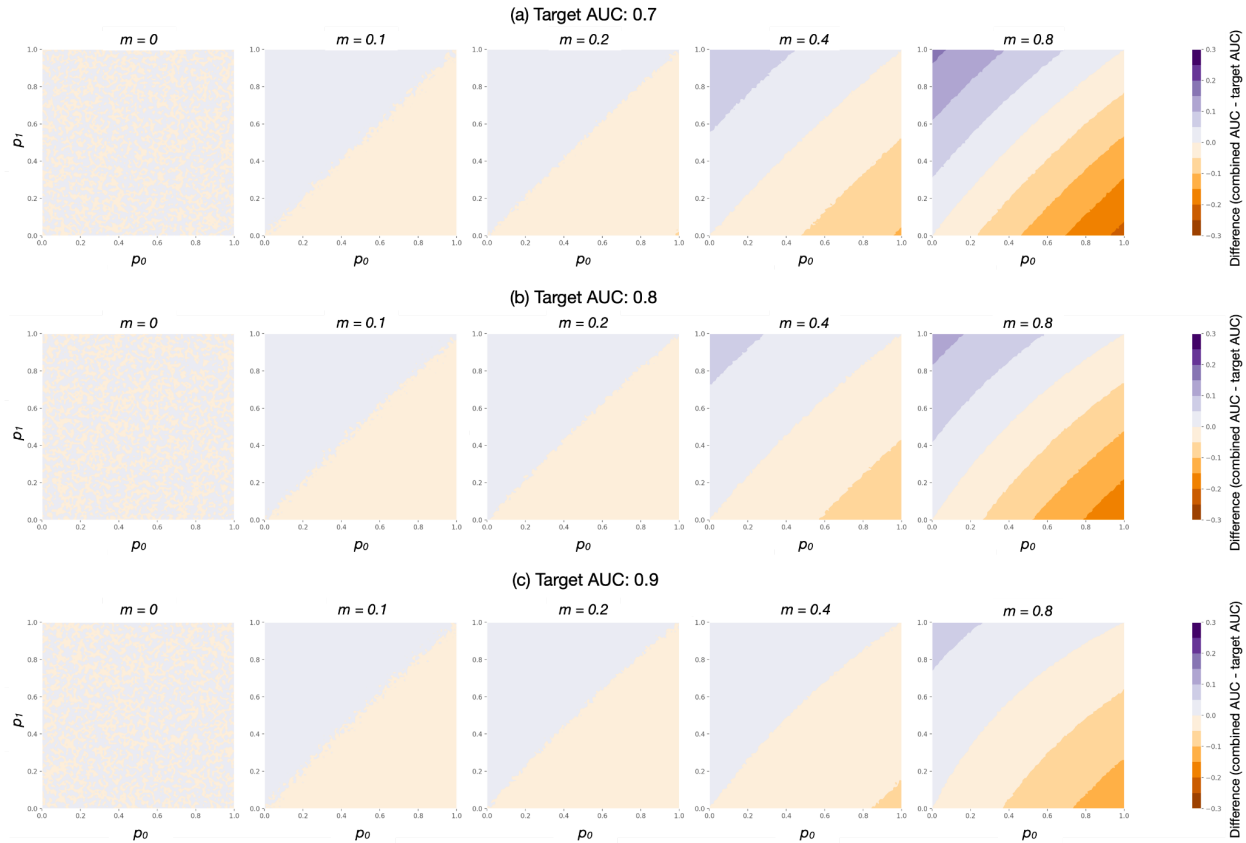

Supplementary Figure 5: Simulation study to measure the difference between combined and target AUC while altering the model bias $m$, the probability of sampling a positive from Set 1 $p_1 = P(S_1 = 1)$, and the probability of sampling a negative from Set 1 $p_0 = P(S_0 = 1)$. The target AUCs are 0.7, 0.8, and 0.9.

For this simulation study, we generated 101 linearly spaced values between 0 and 1 for $p_1$ and $p_0$, chose 5 examples of the model bias $m$, and calculated the constant $a$ for the target AUC of 0.7. For each simulation, we uniformly randomly generated 100 to 10,000 positive and negative cases each and randomly chose which set they belonged to using the Bernoulli distribution. We averaged 100 simulations for each value.

In our simulations (Supplementary Figure 5), we observed that when $m > 0$, increasing $p_1$ and decreasing $p_0$ inflated the combined AUC, while the inverse deflated it. Intuitively, we observed that when the model bias is high, having positives from the biased Set 1 and negatives from the unbiased Set 0 is bias-aligned and inflated the AUC. Having positives from the unbiased Set 0 and negatives from the biased Set 1 is bias-conflicting and deflated the AUC. At low bias $m = 0.1$, the AUC difference is zero when $p_1 \approx p_0$. However, increasing the model bias $m$ or the target AUC weakened this observation, as we observed a curved line. This occurs because the CDF of the first term is asymptotically approaching 1, requiring a higher ratio of bias-aligned samples to achieve the zero AUC difference.

Still, we can mitigate an inflated or deflated AUC by sampling positive and negative samples from the biased set with the same probability. Likewise, we can also maintain the same positive prevalence for both sets to maintain the combined AUC. This finding is crucial for preventing bias when training models (e.g. dataset-balancing) and also designing test datasets from multiple sites, where we can maintain the cancer prevalence for each group.

# Supplementary Section 6: Stratification over Shortcuts

Shortcuts can obscure evaluation over aggregated datasets via the AUC paradox. Section 3.3 examined stratifications at the extremes, i.e. for the bias-aligned and bias-conflicting subsets. We now look at combined AUC in more detail by smoothly adjusting the composition of the evaluation subset. Fixing the evaluation set to the size of the test set, we proportionally adjusted the composition based on the shortcut attribute by sampling with replacement for each attribute value. This method maintained the original cancer prevalence for each attribute value (e.g. WUSTL vs OPTIMAM, screening vs diagnostic, and HS vs SD).

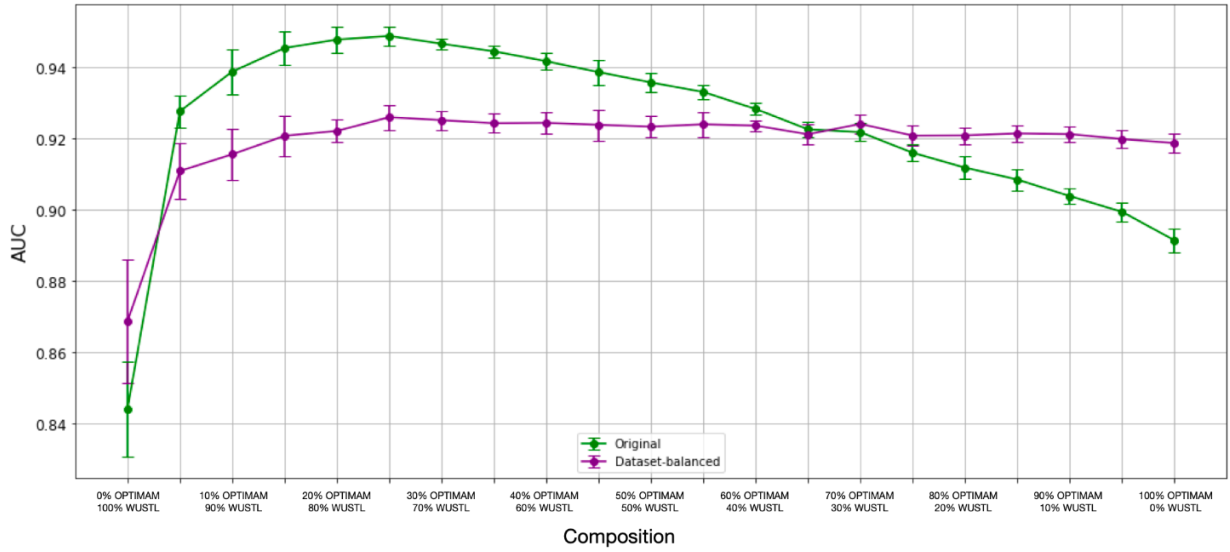

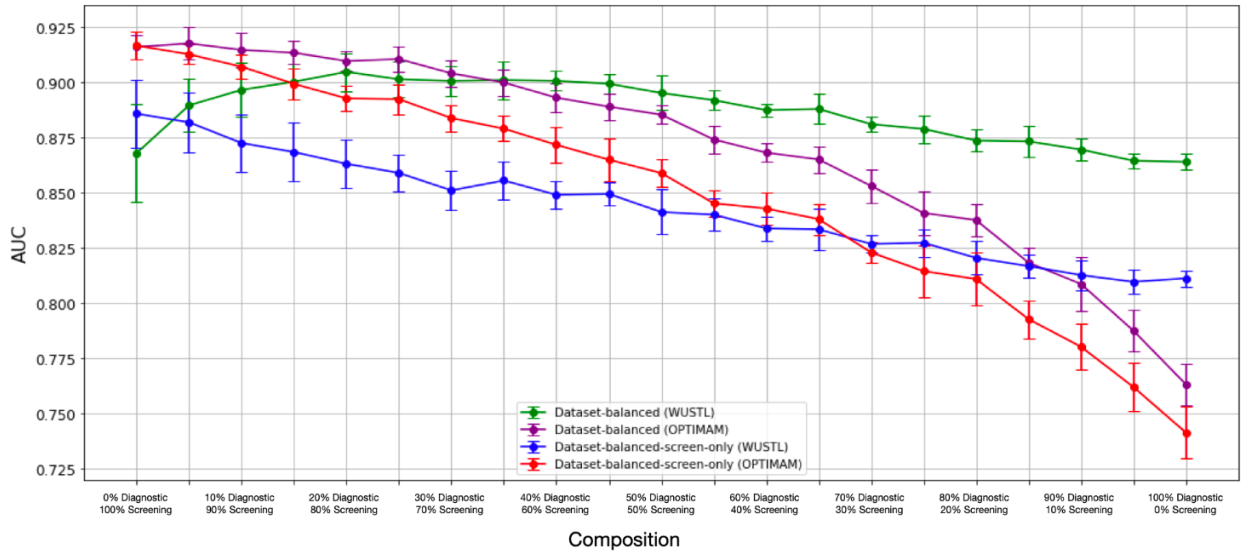

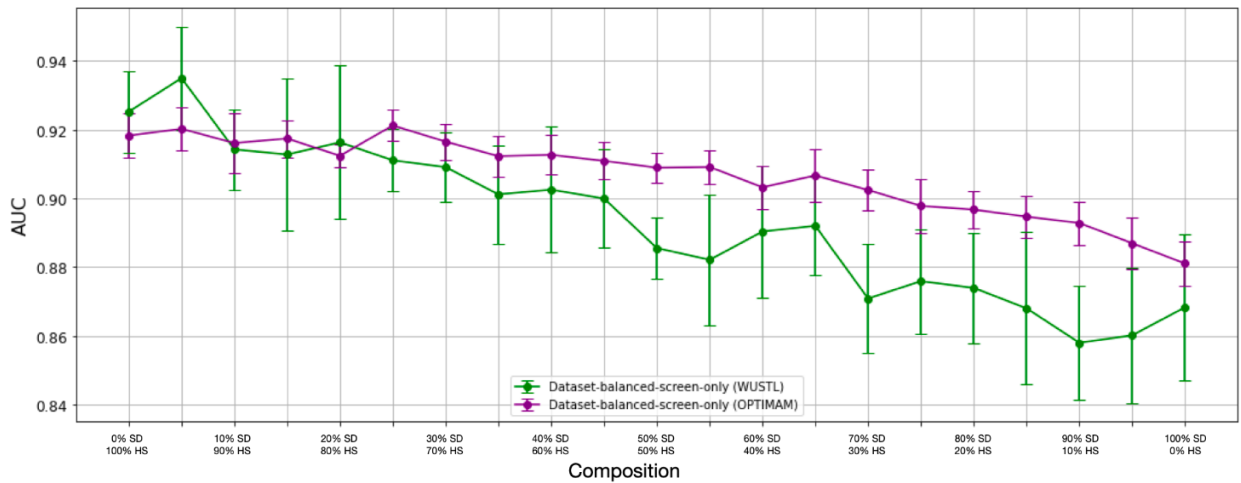

Supplementary Figure 6: Figure (a) contains the AUCs when varying the dataset source composition in the evaluation set. Figure (b) contains AUCs when varying the exam type composition in the evaluation set. Figure (c) contains AUCs when varying the scanner model composition in the evaluation set. Error bars are the standard deviations of the AUCs for 10 random subsets per point.

Supplementary Figure 6 reveals how altering the composition of datasets with the attribute values can change the AUC. Ideally, a model agnostic to shortcuts would form a straight line, but our analysis reveals that the line can be concave, as mixing the attribute values increases performance.

In Supplementary Figure 6a, the original model's curve shows that the introduction of any OPTIMAM data sharply increases performance. This reaction is due to the cancer prevalence of OPTIMAM being 30x that of WUSTL (31.4% vs 0.9%, Figure 1). Increasing the contribution from OPTIMAM dataset increases the number of cancer examples and results in the inflated AUC due to the dataset shortcut in which OPTIMAM exams tend to have higher scores.  Adding more OPTIMAM exams initially increased performance until it reached 30%. This observation occurs because, proportionally, more non-cancer exams from OPTIMAM are added and more WUSTL non-cancer exams are removed from the evaluation set, working against the learned shortcut. The dataset-balancing solution observed a more stable and flatter performance curve as the composition varied. However, remnants of the problem persists, as the sets with both WUSTL and OPTIMAM data have higher AUCs than the individual sets.

Next, stratification by exam type was considered (Supplementary Figure 6b). Diagnostic mammography has a higher cancer prevalence naturally. We observed a similar trend in breast density, as dense breasts have a higher cancer risk than fatty breasts[8]. From this point of view, learning the correlation may be considered a valid signal. The diagnostic subset is considerably smaller than the screening subset in OPTIMAM (243 vs 1880) and has a different cancer prevalence (74.1% vs 31.4%). This ~2.5x difference in prevalence does not appear to be extreme enough to cause a shortcut to be learned, and performance decreased as more diagnostic exams were added. Conversely, we do see the AUC paradox in the WUSTL subset, which has a 10x difference in cancer prevalence over exam type (10.2% vs 0.9%). In WUSTL, we observed that adding diagnostic exams initially increases performance and then decreases performance after reaching 20% diagnostic exams. Training on screening only images removes the concave shape, so no AUC paradox is observed.

Supplementary Figure 6c shows the AUC for the data balanced screen-only model over the scanner composition broken down by dataset. Due to the lack of cancer exams for some subgroups, the standard error bars are wide, leading to difficult assessment of the true impact of the shortcut. In general, the trend is the same across both datasets in that reported performance steadily increases as the composition includes more HS exams, likely due to the fact that the model is trained mostly on HS exams.

In short, large differences in performance can be achieved through slight variations of the dataset. In some cases, how the trend changes as the composition changes can be another method to identify shortcuts. Understanding the aggregate performance is not enough, as it is merely the performance under a specific composition. In the case of innately biased factors such as the exam type, measuring performance isolated to the cohorts separately gives the most meaningful results.